\documentclass{article}
\usepackage{ijcai22}

\usepackage{times}
\usepackage{soul}
\usepackage{url}
\usepackage[hidelinks]{hyperref}
\usepackage[utf8]{inputenc}
\usepackage[small]{caption}
\usepackage{graphicx}
\usepackage{amsmath}
\usepackage{amsthm}
\usepackage{booktabs}
\usepackage{algorithm}
\usepackage{algorithmic}
\usepackage{subfigure}
\usepackage{multirow}
\usepackage{makecell}
\usepackage{xcolor}
\usepackage{pifont}
\usepackage{amsfonts}

\usepackage{bbm}
\usepackage[switch]{lineno}

\newcommand{\tabincell}[2]{\begin{tabular}{@{}#1@{}}#2\end{tabular}}
\newcommand{\ignore}[1]{}

\title{Learning Target-aware Representation for Visual Tracking \\ via Informative Interactions}
\author{
Mingzhe Guo$^1$\thanks{Equal Contribution. ~$^\dagger$ Corresponding author.}\and
Zhipeng Zhang$^2$\footnotemark[1]\and
Heng Fan$^3$\and
Liping Jing$^1$\footnotemark[2]\and\\
Yilin Lyu$^1$\and
Bing Li$^2$\And
Weiming Hu$^2$\\
\affiliations
$^1$Beijing Key Lab of Traffic Data Analysis and Mining, Beijing Jiaotong University\\
$^2$NLPR, Institute of Automation, Chinese Academy of Sciences (CASIA)\\
$^3$Department of Computer Science and Engineering, University of North Texas, Denton, TX USA\\
\emails
\{mingzheguo, yilinlyu, lpjing\}@bjtu.edu.cn, zhangzhipeng2017@ia.ac.cn, \\heng.fan@unt.edu, \{bli, wmhu\}@nlpr.ia.ac.cn
}

\begin{document}

\maketitle

\begin{abstract}
We introduce a novel backbone architecture to improve target-perception ability of feature representation for tracking. Specifically, having observed that de facto frameworks perform feature matching simply using the outputs from backbone for target localization, there is no direct feedback from the matching module to the backbone network, especially the shallow layers. More concretely, only the matching module can \textbf{directly} access the target information (in the reference frame), while the representation learning of candidate frame is blind to the reference target. As a consequence, the accumulation effect of target-irrelevant interference in the shallow stages may degrade the feature quality of deeper layers. In this paper, we approach the problem from a different angle by conducting multiple branch-wise interactions \textbf{in}side the Siamese-like \textbf{b}ackbone \textbf{n}etworks (\textbf{InBN}). At the core of \textbf{InBN} is a general interaction modeler ({\bf GIM}) that injects the prior knowledge of reference image to different stages of the backbone network, leading to better target-perception and robust distractor-resistance of candidate feature representation with negligible computation cost. The proposed \textbf{GIM} module and \textbf{InBN} mechanism are general and applicable to different backbone types including CNN and Transformer for improvements, as evidenced by our extensive experiments on multiple benchmarks. In particular, the CNN version (based on SiamCAR) improves the baseline with 3.2/6.9 absolute gains of SUC on LaSOT/TNL2K, respectively. The Transformer version obtains SUC scores of 65.7/52.0 on LaSOT/TNL2K, which are on par with recent state of the arts. Code and models will be released.
\end{abstract}

\section{Introduction}
As one of the most fundamental tasks in computer vision, visual object tracking (VOT) aims to estimate the trajectory of the designated target in a video sequence. Embracing the powerful deep networks as appearance modeling tools, unprecedented achievement has been witnessed in the past years in tracking community. One representative paradigm, namely two-stream network, has been widely adopted in recent tracking methods, \emph{e.g.,} Siamese and Correlation-Filter (CF) based approaches ~\cite{Siamfc,Siamrpn,SiamDW,Siammask,Atom,Dimp}. The consensus for constructing a robust two-stream tracker is to learn an informative visual representation, or in other words, ``feature matters''~\cite{wangnaiyan,ObjRep}.

\begin{figure}[t]
   \begin{minipage}{0.99\linewidth}
   \centerline{\includegraphics[width=\textwidth]{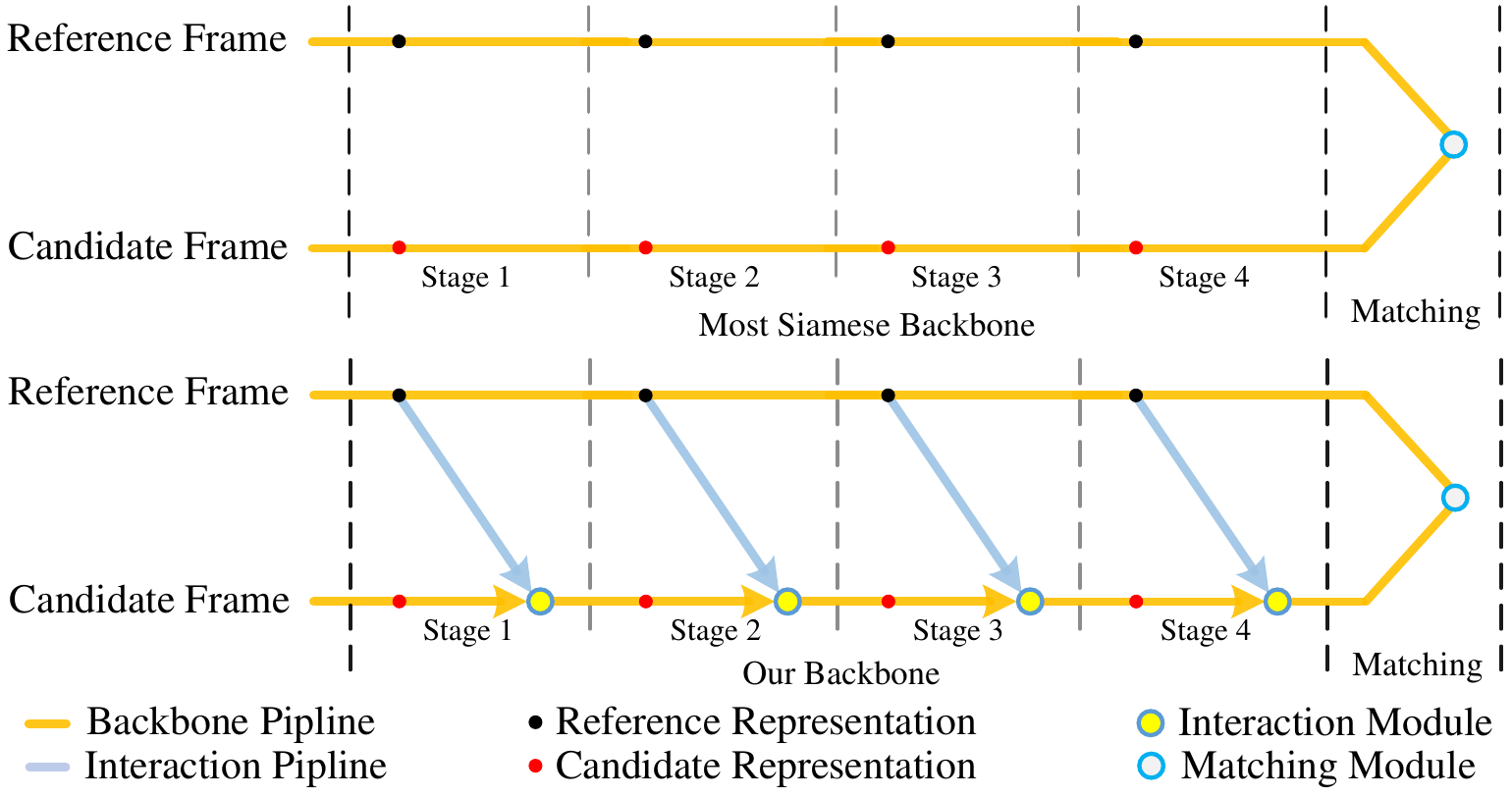}}
   \end{minipage}
   
   \caption{Comparison between two types of backbone networks in tracking. Most Siamese trackers solely use the backbone network to extract representations without interactions (the first row). In contrast, our backbone performs multiple interactions inside different stages of the backbone, which enhances the distractor-aware ability and target-aware representations learning (the second row).}
   \label{fig:backbone}
   \end{figure}

For tracking task, an informative feature requires the target representation clearly distinguishable from background and objects with similar semantics (\emph{i.e.,} distractors). Fortunately, other vision tasks, \emph{e.g.} classification, detection and segmentation, also enjoy sharp contrast between background and objects. A new paradigm from the basic classification task, \emph{i.e., backbone network}, can thus be seamlessly transferred to advance other fields. Object tracking also benefits from this bonus, where different backbone networks are introduced to learn strong visual representation~\cite{SiamDW,Siamrpn++}. But itch of scabies for tracking mostly attributes to surrounding distractors. Besides better feature extractors, a tracking algorithm should also consider how to reinforce target-relevant representation as well as avoiding negative influence from distractors. Previous works, \emph{e.g.,} SiamAttn~\cite{SiamAttn}, try to alleviate this issue by injecting prior knowledge of the reference image (\emph{i.e.,} the first frame) to enhance features of candidate frame. However, only using the target-guided candidate feature for matching without direct feedback to backbone network is not enough. The potential problem is that the branch corresponding to the candidate image cannot well ``sense'' information of the reference target in its feature learning. Besides, the accumulation effect of target-irrelevant interference in the shallow stages may degrade the feature quality of deeper layers. In a nutshell, the under utilization of target information during representation learning may compromise the ability of filtering unrelated distractors, which results in inferior performance.

In this paper, we set out to address the aforementioned issues by proposing a mechanism called ``{\it interaction inside the backbone network} (\textbf{InBN})''. As discussed, the anaemic distractor-aware ability at the early learning stages blames on less exposure to prior information. Therefore, one intuitive but rational solution is to increase the branch-wise interaction frequency inside the backbone network of a two-stream paradigm. In our design, we take inspiration from recent arising vision Transformer architecture~\cite{Vit}, which is adept on modeling global visual relations~\cite{Understanding,Non-local} and can naturally process features from multi-modalities. The proposed general interaction modeler, dubbed as \textbf{GIM}, strengthens the representation of candidate image by learning its relation with reference representation. The relation learning is realized by the context-aware self-attention (CSA) and target-aware cross-attention (TCA) modules (as described later).

For the first time, we integrate the cross-processing modeler into different stages of a backbone network in visual tracking, which can continuously expose prior information to backbone modeling, as shown in Fig.~\ref{fig:backbone}. The \textbf{multiple interactions} pattern improves the distractor-aware ability of the early stages at a backbone network through feature aggregation in CSA and relation learning in TCA. Notably, the proposed method can not only adapt to a CNN network but also work well with a Transformer backbone. When equipping the transformer backbone with the proposed GIM and InBN mechanism, we surprisingly observe that complicated matching networks as in TransT~\cite{TransT} and AutoMatch~\cite{AutoMatch} are not necessary, where a simple cross-correlation can show promising results.

A series of experiments are conducted to prove the generality and effectiveness of the proposed LiBN and GIM. Taking SiamCAR~\cite{SiamCAR} as the baseline tracker, we directly apply GIM and LiBN to its CNN backbone, achieving 3.2 points gains on LaSOT (53.9 vs 50.7). When employing Transformer backbone, our model further improves the performance to SUC of 65.7 on LaSOT.

The main contributions of this work are as follows:

\begin{itemize}
   \item We present the first work demonstrating that information flow between reference and candidate frames can be realized inside the backbone network, which is clearly different from previous feature matching way that does not provide direct feedback to backbones in visual tracking.
   
   \item We prove the effectiveness of the proposed GIM module and InBN mechanism on both CNN and Transformer backbone networks, which makes it a general modeler to improve representation qualities in tracking.
\end{itemize}

\begin{figure*}[t]
   \begin{minipage}{0.99\linewidth}
   \centerline{\includegraphics[width=\textwidth]{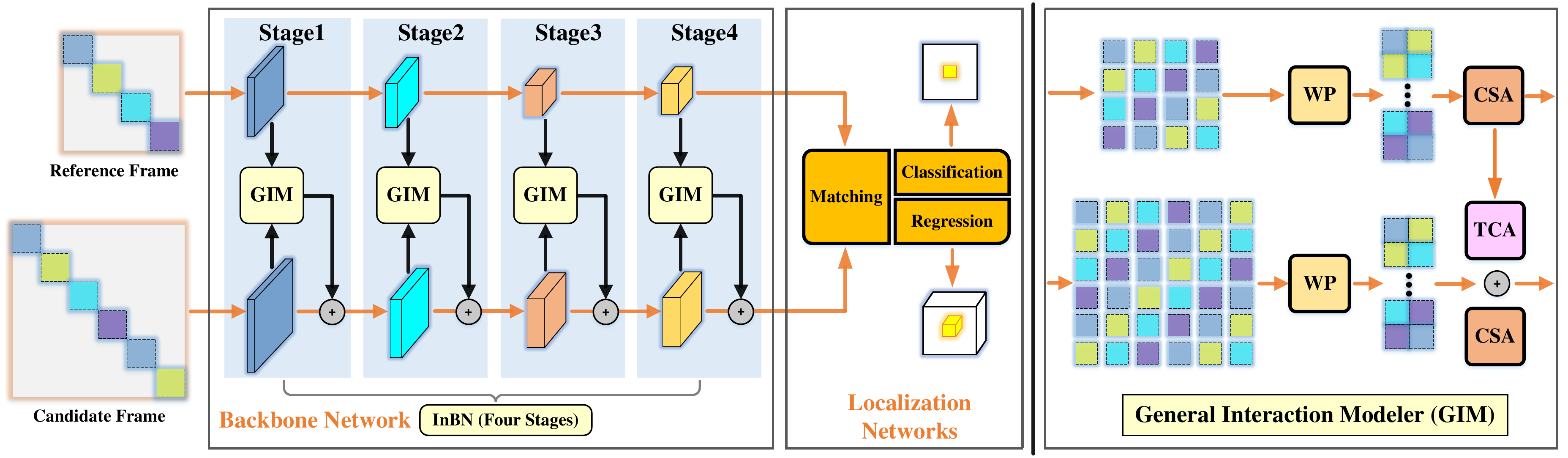}}
   \end{minipage}
   \caption{Architecture of the proposed tracking framework with General Interaction Modeler \textbf{(GIM)}. The window process (\textbf{WP}) ensures the successive attention modules consider both global and local context. The context-aware self-attention \textbf{(CSA)} module and target-aware cross-attention \textbf{(TCA)} module are introduced to improve feature representation learning. Notably, our method can be applied to both CNN backbones like ResNet and Transformer backbones.}
   \label{frame}
   \end{figure*}

\section{Related Work}

{\bf Visual Object Tracking.} The Siamese network has been widely used in visual object tracking in recent years~\cite{Sint,Siamfc,Siamrpn,Siamrpn++,DaSiamRPN,Ocean,SiamDW,TransT,Atom,Dimp}.
SiamFC~\cite{Siamfc} first adopts cross-correlation in the Siamese framework, which tracks a target by template matching. By exploiting the region proposal network (RPN)~\cite{FasterRCNN}, SiamRPN~\cite{Siamrpn} and its variants~\cite{SiamCAR,SiamAttn} achieve more precise target scale estimation and fast speed.
Later, further improvements upon Siamese trackers have been made, including enhancing feature representations by deeper convolutional backbones~\cite{Siamrpn++,SiamDW}, designing more practical update mechanisms~\cite{MDNet,UpdateNet,STMTracker}, introducing anchor-free regression to target estimation~\cite{SiamCAR,Ocean}, replacing cross-correlation with the automatically learned matching networks~\cite{AutoMatch}, introducing vision Transformer~\cite{TransT}, and so on. Another important branch, \emph{i.e.,} online trackers, always construct models with discriminative correlation filter~\cite{ECO,Atom,Dimp,Prdimp}. 
Despite of their success, only relying on a sole backbone to extract image representation without any interactive learning may introduce distractors and limit the tracking performance. In this work, we introduce a general interaction modeler to alleviate this issue.

\noindent
{\bf Vision Transformer.} Transformer is originally proposed in ~\cite{Transformer} for the task of machine translation. Recently it has been introduced into vision tasks and shows great potential~\cite{Vit,ViT2T,PVT,BoNT,LVT}. The core idea of Transformer is self-attention mechanism, which takes a sequence as the input, and builds similarity scores of each two feature vectors as attention weights to reinforce or suppress visual representations of corresponding positions. ViT~\cite{Vit} adopts a convolution-free transformer architecture for image classification, which processes the input image into a sequence of patch tokens. In addition to classification, Transformer is also used in other vision tasks and achieves comparable performances, such as object detection~\cite{Detr,DeforDetr}, semantic segmentation~\cite{Segmenter}, multiple object tracking~\cite{TransTrack,Trackformer}, etc. In visual tracking, TrDiMP~\cite{TrDiMP} uses transformer to learn temporal relations between frames, and TransT~\cite{TransT} employs cross-attention in matching process to enhance semantic messages. 
Transformer is naturally suitable for processing multi-modality features. Surprisingly, no prior works in visual tracking consider building interactions inside the two-stream backbone network. The information in the reference image (or template image named in some works) is crucial to explore more target-relevant clues and track targets in complex scenes, especially in the presence of similar distractors. Our work aims to narrow these gaps by proposing an general interaction modeler which promotes the representation of candidate image at different stages of the backbone network.

\section{Tracking with General Interaction Modeler}
\label{sec:InMo}

In this section, we elaborate on the proposed General Interaction Modeler (\textbf{GIM}) and its integration to different backbone networks via the proposed \textbf{InBN} mechanism for improving representation learning in visual tracking.

\subsection{General Interaction Modeler}
General Interaction Modeler (GIM) contains three essential components including window process (WP), context-aware self-attention (CSA) module and target-aware cross-attention (TCA) module, as illustrated in Fig.~\ref{frame}. The WP module is designed to decrease computational complexity and meanwhile increase the local perception ability. The CSA module aggregates context information by global attention modeling. The TCA module enhances target-related features and suppresses distractor responses by injecting prior knowledge of the reference image to representation learning. In the following, we will detail each module one by one.

\subsubsection{Local Perception via Window Process (WP)} 
\label{WP}
Our GIM is built upon the Transformer-like structure as described above. The Achilles' Heel is its quadratic cost for global attention learning and lack of local perception ability~\cite{swin_trans,twins_trans}. Motivated by recent vision transformers~\cite{swin_trans}, we design the window process to merge feature vectors of non-overlapping local areas before attention modules.

Window process (WP) first partitions the input feature $\mathbf{f} \in \mathbb{R}^{B \times H \times W \times C}$ 
($B$ is the batch size) by performing non-overlapping $win \times win$ windows ($win=7$) on it. Then each group containing $win^2$ feature vectors is regarded as a modeling unit and forms a new dimension with size of $\mathbf{f}_{group} \in \mathbb{R}^{B \times win^2 \times \frac{H}{win} \times \frac{W}{win} \times C}$.
To fit the requirement of following attention modules, we flatten the spatial dimension of $\mathbf{f}_{group}$ to $\mathbf{f}_{WP} \in \mathbb{R}^{B \times win^2 \times \frac{HW}{win^2} \times C}$, whose sequence length is $\frac{HW}{win^2}$. We perform attention on dimensions of $\frac{HW}{win^2}$ and $C$ instead of $win^2$ and $C$ as in Swin-Trans~\cite{swin_trans}, which ensures the feasibility to compute the cross-attention weights $\mathbf{W}_{attn} \in \mathbb{R}^{B \times win^2 \times \frac{{H_x}{W_x}}{win^2} \times \frac{{H_z}{W_z}}{win^2}}$ with Einstein Summation Convention~\cite{einsum} ($z$ and $x$ represent the reference and candidate images respectively). By regarding each window as a computing unit and then modeling relations between different windows, local clues are used without affecting global receptive fields.

\subsubsection{Context-aware Self-attention (CSA)}
The proposed CSA aggregates context information by performing multi-head self-attention (MHSA) on the partitioned input feature $\mathbf{f}_{WP}$. In particular, given an input feature sequence $\mathbf{f}_{seq} \in \mathbb{R}^{B \times HW \times C}$, the inner single-head attention function first maps it into query $\mathbf{Q} \in \mathbb{R}^{B \times N_{Q} \times d}$, key $\mathbf{K} \in \mathbb{R}^{B \times N_{K} \times d}$ 
and value $\mathbf{V} \in \mathbb{R}^{B \times N_{V} \times d}$ ($N_Q$, $N_K$ and $N_V$ are sequence lengths) with mapping weights ${\bf{W}}_{Q},{\bf{W}}_{K},{\bf{W}}_{V} \in \mathbb{R}^{C \times d}$, then performs dot-product on them, 
\begin{equation}
    \begin{aligned}
        \mathbf{Q}={\mathbf{f}_{seq}}{\bf{W}}_{Q},
        \mathbf{K}={\mathbf{f}_{seq}}{\bf{W}}_{K},
        \mathbf{V}={\mathbf{f}_{seq}}{\bf{W}}_{V},\\
      {\rm{Attention}}(\mathbf{Q},\mathbf{K},\mathbf{V}) = {\rm{softmax}}(\frac{\mathbf{Q}\mathbf{K}^\top}{\sqrt{d}})\mathbf{V}, 
    \end{aligned}
\label{eq-s-att}
\end{equation}
where $d$ is the number of channels. In our model, we directly split the features along channel dimension for multiple heads, which is computed as follows,

\begin{equation}
\label{eq-m-att}
\begin{aligned}
   {\rm{MultiHead}}(\mathbf{Q},\mathbf{K},\mathbf{V}) &= {\rm{Concat}}({\mathbf{H}_1},...,{\mathbf{H}_{{n}}}){\mathbf{W}_{map}}, \\
   {\mathbf{H}_i} = &\,{\rm{Attention}}(\mathbf{Q}_i,\mathbf{K}_i,\mathbf{V}_i),\\
   \mathbf{Q} = &\,{\rm{Concat}}({\mathbf{Q}_1},...,{\mathbf{Q}_{n}}),\\
   \mathbf{K} = &\,{\rm{Concat}}({\mathbf{K}_1},...,{\mathbf{K}_{n}}),\\
   \mathbf{V} = &\,{\rm{Concat}}({\mathbf{V}_1},...,{\mathbf{V}_{n}}), 
\end{aligned}
\end{equation}
where $\mathbf{Q}_i \in \mathbb{R}^{B \times N_Q \times \frac{d}{n}}$, $\mathbf{K}_i \in \mathbb{R}^{B \times N_K \times \frac{d}{n}}$, 
$\mathbf{V}_i \in \mathbb{R}^{B \times N_V \times \frac{d}{n}}$ are divided parts, 
and $\mathbf{W}_{map} \in \mathbb{R}^{d \times C}$ is mapping weights. 
We employ $n=4, C=256,$ and $d=256$ to achieve real-time tracking speed without other specified.

CSAs are applied on features of both reference and candidate images.
It computes similarity scores between each two positions and 
uses the scores as attention weights to aggregate context visual features. By considering $\mathbf{f}_{WP}$ as a sequence, the sequence length after WP is $\frac{HW}{win^2}$ instead of original $HW$, which has linear cost and promotes modeling speed. In a nutshell, the mechanism of CSA for the input $\mathbf{f}_{WP} \in \mathbb{R}^{B \times win^2 \times \frac{HW}{win^2} \times C}$ can be summarized as

\begin{equation}
   \label{equation:TFA}
\begin{aligned}
   \mathbf{Q}={\mathbf{f}_{WP}}{\bf{W}}_{Q},
   \mathbf{K}={\mathbf{f}_{WP}}{\bf{W}}_{K},
   \mathbf{V}={\mathbf{f}_{WP}}{\bf{W}}_{V},\\
   {{\mathbf{f}}_{CSA}}={{\mathbf{f}_{WP}}+\rm{MultiHead}}(\mathbf{Q},\mathbf{K},\mathbf{V}),
\end{aligned}
\end{equation}
where ${\bf{W}}_{Q},{\bf{W}}_{K},{\bf{W}}_{V} \in \mathbb{R}^{C \times d}$ are mapping weights, 
$\mathbf{f}_{CSA} \in \mathbb{R}^{B \times win^2 \times \frac{HW}{win^2} \times C}$ is the output of CSA.

\subsubsection{Target-aware Cross-attention (TCA)}
As mentioned before, Siamese tracking is conducted on a two-stream network, which is different from the typical one-stream classification model. Therefore, it is reasonable to consider the interaction between these two branches during backbone design. In this work, we propose the target-aware cross-attention (TCA) to inject the reference information to the branch of the candidate image, which guides the backbone to perceive the target and filter irrelevant distractors.

For implementation, TCA takes two window-processed features ($\mathbf{f}_{x_{WP}} \in \mathbb{R}^{B \times win^2 \times \frac{{H_x}{W_x}}{win^2}  \times C}$ for candidate image and $\mathbf{f}_{z_{WP}} \in \mathbb{R}^{B \times win^2 \times \frac{{H_z}{W_z}}{win^2} \times C}$ for reference image) as inputs and maps them to $\mathbf{Q}$ and $\mathbf{K},\mathbf{V}$ respectively. Through multi-head attention, the contents in $\mathbf{f}_{x_{WP}}$ are refined by the prior knowledge in $\mathbf{f}_{z_{WP}}$, where the target-related ones are strengthened and the irrelevant ones are suppressed. TCA is formulated as,

\begin{equation}
   \label{equation:IGA}
\begin{aligned}
   \mathbf{Q}=&\mathbf{f}_{x_{WP}}{\bf{W}}_{Q},
   \mathbf{K}=\mathbf{f}_{z_{WP}}{\bf{W}}_{K},
   \mathbf{V}=\mathbf{f}_{z_{WP}}{\bf{W}}_{V},\\
   &{{\mathbf{f}}_{TCA}}={\mathbf{f}_{x_{WP}}+\rm{MultiHead}}(\mathbf{Q},\mathbf{K},\mathbf{V}),
\end{aligned}
\end{equation}
where ${\bf{W}}_{Q},{\bf{W}}_{K},{\bf{W}}_{V} \in \mathbb{R}^{C \times d}$ are mapping weights, 
$\mathbf{f}_{TCA} \in \mathbb{R}^{B \times win^2 \times \frac{{H_x}{W_x}}{win^2}  \times C}$ is the output of TCA.

Here we analyze how WP could make the computation of attention learning linear complexity. 
Without loss of generality, we assume $H \% win = 0$ and $W \% win = 0$. 
Taking each window as the computing unit, the sequence length is $\frac{HW}{win^2}$. 
CSA or TCA first computes relations between feature vectors of the same position in different windows, and the computation cost is $\mathcal{O} (\frac{H^2W^2}{win^4}d)$. 
Then the total cost for $win^2$ positions is $\mathcal{O} (\frac{H^2W^2}{win^2}d)$. 
If we let $k_1=\frac{H}{win}$ and $k_2 = \frac{W}{win}$, the cost can be computed 
as $\mathcal{O} (k_1k_2HWd)$, which is significantly more efficient when $k_1 \ll H$ and $k_2 \ll W$ and grows linearly with $HW$ if $k_1$ and $k_2$ are fixed.

\begin{table}[t]
   \tabcolsep 1pt
   \caption{Configurations of the built backbone with GIM and InBN.}
   \label{tab: backbone}
   \centering 
   \resizebox{0.48\textwidth}{!}{
   \begin{tabular}{c|c|c|c|c|c|cc}
      \toprule
      &\multicolumn{3}{c|}{CNN Backbone} &\multicolumn{3}{c}{Transformer Backbone} &\\
      \cline{2-7}
      \specialrule{0em}{1pt}{1pt}
      & Output Size & Layer Name &Parameter & Output Size & Layer Name &Parameter & \\
      \midrule
      
      \multirow{2}{*}[-2.5ex]{Stage 1} & \multirow{2}{*}[-2.5ex]{\scalebox{1.3}{$\frac{H}{4}\times \frac{W}{4}$}} & \tabincell{c}{Convolution\\Block} &\tabincell{c}{$P_1=4$; $C_1=256$} & \multirow{2}{*}[-2.5ex]{\scalebox{1.3}{$\frac{H}{4}\times \frac{W}{4}$}} & \tabincell{c}{Patch\\Embedding} & $P_1=4$; $C_1=96$\\
      \cline{3-4}
      \cline{6-7}
      \specialrule{0em}{1pt}{1pt}
      & & \tabincell{c}{GIM} & 
      $\begin{bmatrix}
         \begin{array}{c}
            WP \\
            CSA \\
            (TCA) \\
         \end{array}
      \end{bmatrix} \times 1$ 
      & & \tabincell{c}{GIM} & 
      $\begin{bmatrix}
         \begin{array}{c}
            WP \\
            CSA \\
            (TCA) \\
         \end{array}
      \end{bmatrix} \times 2$ \\
      \specialrule{0em}{1pt}{1pt}
      \midrule
      
      \multirow{2}{*}[-2.5ex]{Stage 2} & \multirow{2}{*}[-2.5ex]{\scalebox{1.3}{$\frac{H}{8}\times \frac{W}{8}$}} & \tabincell{c}{Convolution\\Block} &\tabincell{c}{$P_2=2$; $C_2=512$} & \multirow{2}{*}[-2.5ex]{\scalebox{1.3}{$\frac{H}{8}\times \frac{W}{8}$}} & \tabincell{c}{Patch\\Embedding} & $P_2=2$; $C_2=192$\\
      \cline{3-4}
      \cline{6-7}
      \specialrule{0em}{1pt}{1pt}
      & & \tabincell{c}{GIM} & 
      $\begin{bmatrix}
         \begin{array}{c}
            WP \\
            CSA \\
            (TCA) \\
         \end{array}
      \end{bmatrix} \times 1$ 
      & & \tabincell{c}{GIM} & 
      $\begin{bmatrix}
         \begin{array}{c}
            WP \\
            CSA \\
            (TCA) \\
         \end{array}
      \end{bmatrix} \times 2$ \\
      \specialrule{0em}{1pt}{1pt}
      \midrule
      
      \multirow{2}{*}[-2.5ex]{Stage 3} & \multirow{2}{*}[-2.5ex]{\scalebox{1.3}{$\frac{H}{8}\times \frac{W}{8}$}} & \tabincell{c}{Convolution\\Block} &\tabincell{c}{$P_3=1$; $C_3=1024$} & \multirow{2}{*}[-2.5ex]{\scalebox{1.3}{$\frac{H}{16}\times \frac{W}{16}$}} & \tabincell{c}{Patch\\Embedding} & $P_3=2$; $C_3=384$\\
      \cline{3-4}
      \cline{6-7}
      \specialrule{0em}{1pt}{1pt}
      & & \tabincell{c}{GIM} & 
      $\begin{bmatrix}
         \begin{array}{c}
            WP \\
            CSA \\
            (TCA) \\
         \end{array}
      \end{bmatrix} \times 1$ 
      & & \tabincell{c}{GIM} & 
      $\begin{bmatrix}
         \begin{array}{c}
            WP \\
            CSA \\
            (TCA) \\
         \end{array}
      \end{bmatrix} \times 6$ \\
      \specialrule{0em}{1pt}{1pt}
      \midrule
      
      \multirow{2}{*}[-2.5ex]{Stage 4} & \multirow{2}{*}[-2.5ex]{\scalebox{1.3}{$\frac{H}{8}\times \frac{W}{8}$}} & \tabincell{c}{Convolution\\Block} &\tabincell{c}{$P_4=1$; $C_4=2048$} & \multirow{2}{*}[-2.5ex]{\scalebox{1.3}{$\frac{H}{16}\times \frac{W}{16}$}} & \tabincell{c}{Patch\\Embedding} & $P_4=1$; $C_4=768$\\
      \cline{3-4}
      \cline{6-7}
      \specialrule{0em}{1pt}{1pt}
      & & \tabincell{c}{GIM} & 
      $\begin{bmatrix}
         \begin{array}{c}
            WP \\
            CSA \\
            (TCA) \\
         \end{array}
      \end{bmatrix} \times 1$ 
      & & \tabincell{c}{GIM} & 
      $\begin{bmatrix}
         \begin{array}{c}
            WP \\
            CSA \\
            (TCA) \\
         \end{array}
      \end{bmatrix} \times 2$ \\
      \specialrule{0em}{1pt}{1pt}
      \midrule
      
      Output & \scalebox{1.3}{$\frac{H}{8}\times \frac{W}{8}$} & \tabincell{c}{Mapping} &$C=256$ & \scalebox{1.3}{$\frac{H}{8}\times \frac{W}{8}$} & \tabincell{c}{Fusion} &$C=256$ \\
      \bottomrule
   \end{tabular}}
\end{table}

\subsection{Building Backbone Network}
\label{buildingBN}
To demonstrate the generality and effectiveness of the proposed GIM, we follow the InBN mechanism and build two types of backbone networks for Siamese tracking, \textit{i.e.,} CNN and Transformer as shown in Tab.~\ref{tab: backbone}. To unleash the target-perception ability of shallow layers, we adopt a hierarchical design to employ the GIM, which is named as InBN mechanism. We arrange the proposed GIM to different stages of a tracking backbone to conduct \textbf{multiple interactions}. Based on the reference presentation of different resolutions, the network knows what target is the one being tracked, which eventually enhances target-related messages and suppresses distractors hierarchically.

For the CNN-based model, we take the ResNet-50 in SiamCAR~\cite{SiamCAR} as the basic CNN backbone. Following the InBN paradigm, we modify it by directly applying the GIM after each stage. The mapping layers are attached at each stage to unify the output feature dimensions to $256$ (see more details in \cite{SiamCAR}).

\begin{figure}[t]
  \begin{minipage}{0.99\linewidth}
  \centerline{\includegraphics[width=\textwidth]{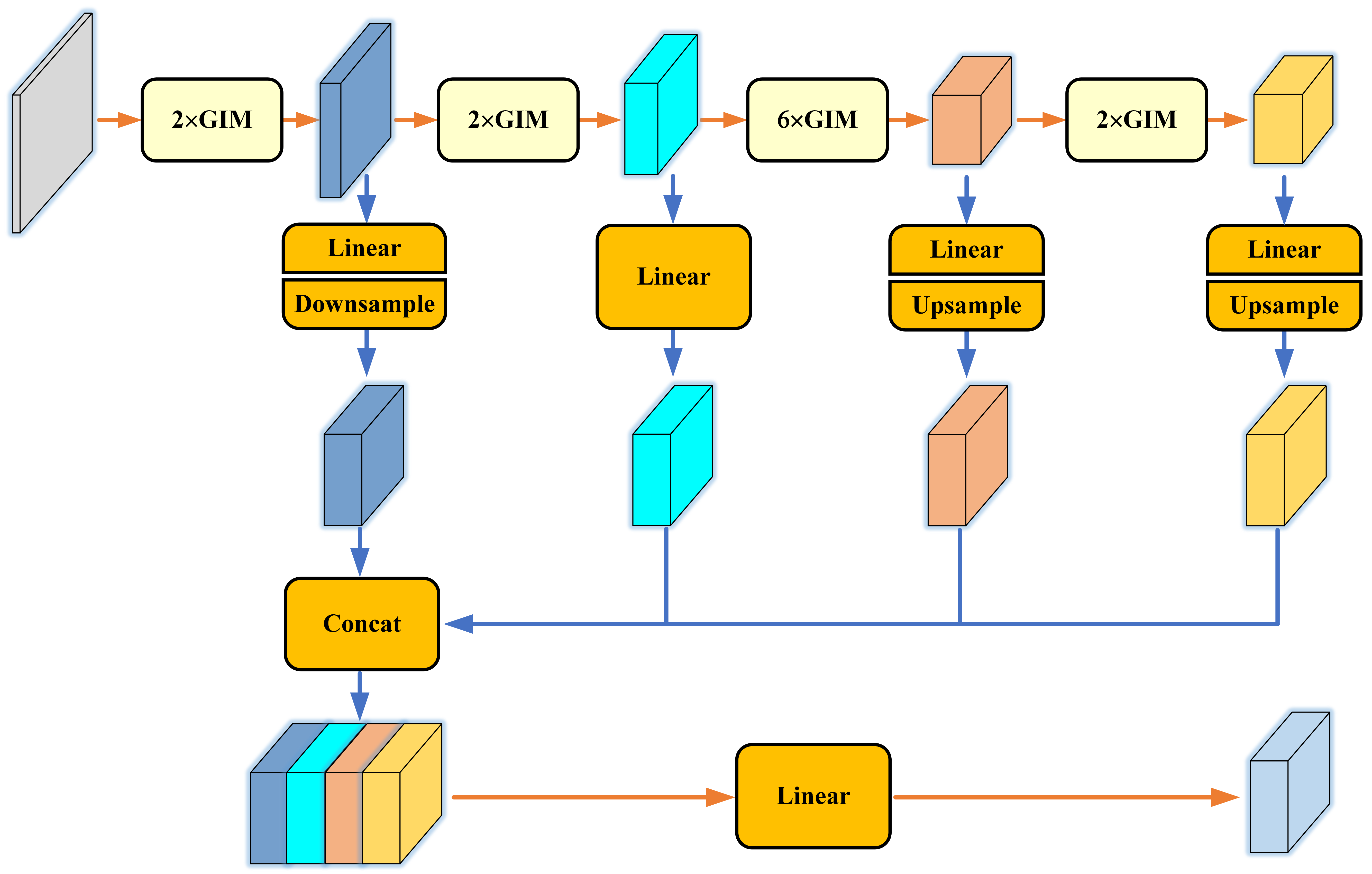}}
  \end{minipage}
   
  \caption{Architecture of fusing multi-stage features in Transformer-based backbone.}
  \label{fig:fusion}
  \end{figure}

For the Transformer-based model, we refer to the structure of Swin-Trans~\cite{swin_trans} to build two-stream backbone by stacking the GIMs. As illustrated in Tab.~\ref{tab: backbone}, the backbone network is divided into four stages, which contain 2, 2, 6, 2 GIMs, respectively. Notably, we only use the TCA module in the last GIM of each stage, which avoids bringing high computation costs. The InBN mechanism allows the backbone to enhance target-relevant presentation gradually by hierarchical interactions. To exploit hierarchical features in the backbone network as the baseline tracker SiamCAR, we also design similar fusion module for TransInMo and TransInMo*. As illustrated in Fig.~\ref{fig:fusion}, for the output features of each stage, we first map their channel dimensions into the same number of $256$ with a $1\times 1$ linear layer. We upsample or downsample these features to the same resolution to align the spatial dimension. Finally, we concatenate all features along channel dimension and use a $1\times 1$ linear transformation to shrink the channel number to $256$.

\begin{table*}[!t]
   \caption{Comparisons on LaSOT, TNL2K, UAV123 and NFS. The best three results 
   are shown in \textbf{\textcolor{red}{red}}, \textbf{\textcolor{green}{green}} and \textbf{\textcolor{blue}{blue}} fonts, respectively.}
   
   \label{tab-sota}
   \begin{center}
   \resizebox{\linewidth}{!}{
   \begin{tabular}{|l|c|ccc|ccc|cc|cc|c|}
   \hline
   \multirow{2}{*}{Method} & \multirow{2}{*}{Source} &\multicolumn{3}{c|}{LaSOT}	
   &\multicolumn{3}{c|}{TNL2K}	&\multicolumn{2}{c|}{UAV123} &\multicolumn{2}{c|}{NFS} & \multirow{2}{*}{FPS} \\
  \cline{3-12}
   & &AUC	&P$_{Norm}$	&P	&AUC &P$_{Norm}$ &P	&AUC &P &AUC &P &\\
   \hline
   \textbf{TransInMo*}	&\textbf{Ours}	&\textcolor{red}{\textbf{65.7}}	&\textcolor{red}{\textbf{76.0}}	&\textcolor{red}{\textbf{70.7}}	
   &\textcolor{red}{\textbf{52.0}}	&\textcolor{red}{\textbf{58.5}} &\textcolor{red}{\textbf{52.7}}
   &\textcolor{red}{\textbf{69.0}}	&\textcolor{green}{\textbf{89.0}}	
   &\textcolor{red}{\textbf{66.8}}	&\textcolor{green}{\textbf{80.2}} &34\\
   \textbf{TransInMo} &\textbf{Ours}	&\textcolor{green}{\textbf{65.3}}	&\textcolor{green}{\textbf{74.6}}	&\textcolor{green}{\textbf{69.9}}	
   &\textcolor{green}{\textbf{51.5}}	&\textcolor{green}{\textbf{57.8}} &\textcolor{green}{\textbf{52.6}}
   &\textcolor{green}{\textbf{68.2}}	&\textcolor{red}{\textbf{90.5}}	
   &\textcolor{green}{\textbf{66.4}}	&\textcolor{red}{\textbf{80.7}} &\textbf{67}\\
   
   TransT	&CVPR2021	&\textcolor{blue}{\textbf{64.9}}	&\textcolor{blue}{\textbf{73.8}}	&\textcolor{blue}{\textbf{69.0}}	
   &\textcolor{blue}{\textbf{50.7}}	&\textcolor{blue}{\textbf{57.1}} &\textcolor{blue}{\textbf{51.7}}
   &\textcolor{blue}{\textbf{68.1}}	&\textcolor{blue}{\textbf{87.6}}	
   &65.7	&\textcolor{blue}{\textbf{78.8}} &50\\
   TrDiMP	&CVPR2021	&63.9	&-	&61.4
   &- &- &- 
   &67.5	&87.2	
   &\textcolor{blue}{\textbf{66.5}}	&78.4 &26\\
   \cline{1-13}
   \textbf{CNNInMo}	&\textbf{Ours}	  &53.9	&61.6	&53.9
   &42.2	&48.6	&41.9	
   &62.9	&81.8 
   &56.0	&68.5 &47\\
   SiamCAR	&CVPR2020	  &50.7	&60.0	&51.0
   &35.3	&43.6	&38.4	
   &61.4	&76.0 
   &52.9	&65.7 &52\\
   \cline{1-13}
   AutoMatch	&ICCV2021	&58.3	&-	&59.9
   &47.2 &- &43.5 
   &63.9	&-	
   &60.6	&- &50\\
   Ocean	&ECCV2020	&56.0	&65.1	&56.6	
   &38.4	&45.4	&37.7	
   &62.1	&-
   &55.3	&- &58\\
   KYS	    &ECCV2020	&55.4	&63.3	&-	
   &44.9	&51.1	&43.5	
   &-	&-	
   &63.5	&77.0 &20\\
   SiamFC++	&AAAI2020	&54.4	&62.3	&54.7	
   &38.6	&45.0	&36.9	
   &63.1	&76.9	
   &58.1	&- &90\\
   PrDiMP &CVPR2020	&59.8	&68.8	&60.8	
   &47.0	&54.0	&45.9	
   &66.8	&87.6	
   &63.5	&76.5 &30\\
   CGACD	&CVPR2020	&51.8	&62.6	&53.5	
   &-	&-	&-	
   &63.3	&83.3	&-	&- &70\\
   SiamAttn	&CVPR2020	&56.0	&64.8	&-	
   &-	&-	&-	
   &65.0	&84.5	
   &-	&-  &45\\
   
   SiamBAN	&CVPR2020	&51.4	&59.8	&52.1	
   &41.0	&48.5	&41.7	
   &63.1	&83.3	
   &59.4	&70.0 &40\\
   DiMP	    &ICCV2019	&56.9	&65.0	&56.7	
   &44.7	&51.3	&43.4	
   &65.4	&85.6	
   &62.7	&75.1 &40\\
   SiamPRN++ &CVPR2019	&49.6	&56.9	&49.1	
   &41.3	&48.2	&41.2	
   &61.3	&80.7	
   &50.8	&50.9 &35\\
   ATOM	    &CVPR2019	&51.5	&57.6	&50.5	
   &40.1	&46.5	&39.2	
   &65.0	&85.8	
   &59.0	&69.4 &30\\
   SiamPRN &CVPR2018	&49.6	&56.9	&49.1	
   &32.9	&36.2	&28.1	
   &55.7	&71.0	
   &48.8	&56.7 &160\\
   ECO	 &ICCV2017	&32.4	&33.8	&30.1	
   &32.6	&37.7	&31.7	
   &52.5	&68.8	
   &46.6	&54.7 &60\\
   SiamFC	&ECCV2016	&33.6	&42.0	&33.9	
   &29.5	&45.0	&28.6	
   &48.5	&64.8	
   &-	&- &58\\
   \hline
   \end{tabular}}
   \end{center}
   \end{table*}

\subsection{Comparison with Other Tracking Backbones}
To our best knowledge, this is the first tracking framework that considers branch-wise interactions inside the backbone network for better feature representation. As shown in Fig.~\ref{fig:backbone}, we categorize existing backbone networks in visual tracking into two types. The first one (see the first row in Fig.~\ref{fig:backbone}), which is employed in the most trackers, simply uses a backbone to extract feature representation and then performs matching without direct feedback for tracking. The lack of reference information makes it hard to robustly locate the target object. Differently, equipping with GIM with InBN mechanism, our backbone network (see the second row in Fig.~\ref{fig:backbone}) is able to not only perform multiple interactions to bridge the target information to candidate frame during feature learning, but also exploit global modeling to enrich target-related representation, resulting in improved feature representations in all layers for better performance, yet with negligible computation.

\section{Experiments}
\label{sec-exp}


\subsection{Implementation Details}
For CNN-based tracker, we apply the proposed \textbf{In}teraction \textbf{Mo}deler to the baseline tracker SiamCAR~\cite{SiamCAR} (named as \textbf{CNNInMo}). For transformer-based tracker (named as \textbf{TransInMo}), we build backbone network following the structions in Sec.~\ref{buildingBN}. A simple depth-wise correlation layer is used for feature matching. Two three-layer MLPs are respectively conducted for classification and regression heads, as similar in~\cite{TransT}. Moreover, we test the influence of complicated matching process by equipping TransInMo with the matching module from TransT~\cite{TransT} (named as \textbf{TransInMo*}). The training settings for CNNInMo follow the baseline tracker SiamCAR (we recommend the readers to \cite{SiamCAR} for more details). We train TransInMo and TransInMo* on the training splits of LaSOT~\cite{LaSOT}, TrackingNet~\cite{trackingnet} and GOT-10k~\cite{GOT10K} and COCO~\cite{COCO} for 1000 epochs (1000 iterations with batch size of 80 per epoch). The AdamW~\cite{AdamW} is used for network optimization, with an initial learning rate of $5e-5$ for the transformer backbone and $5e-4$ for other layers. We decrease the learning rate by a factor of 0.1 for each 400 epochs.

{\noindent \textbf{Loss Functions for Offline Training. }} We detail the loss functions used to train TransInMo and TransInMo* (CNNInMo follows the training process of its baseline tracker SiamCAR~\cite{SiamCAR}).
For classification sub-task, the standard binary cross-entropy loss is employed.
We select the feature points within the ground-truth bounding box as positive samples and the rest are considered as negative samples.
To balance the numbers of positive and negative samples, we set the balance weight of negative 
samples to $\frac{1}{16}$. 
Then the classification loss is defined as
\begin{equation}
\begin{split}
\label{equation:BCE_loss}
\mathcal{L}_{cls} = -\sum_j[y_j{\rm log}(p_j)+(1-y_j){\rm log}(1-p_j)], 
\end{split}
\end{equation}
where $y_j$ is the ground-truth label and $p_j$ is the predicted foreground confidence.
For the regression sub-task, we employ a linear combination of $\ell_1$-norm loss $\mathcal{L}_{1}(.,.)$ 
and the generalized IoU loss $\mathcal{L}_{GIoU}(.,.)$~\cite{GIoU}. 
Different to the classification network, only positive samples contribute to the optimization of regression task. 
The regression loss can be formulated as
\begin{equation}
\begin{split}
\label{equation:bbox_loss}
\mathcal{L}_{reg} = \sum_j\mathbbm{1}_{\{y_j=1\}}[\lambda_{G}\mathcal{L}_{GIoU}(b_j,\hat{b})+\lambda_1\mathcal{L}_1(b_j,\hat{b})]
\end{split}
\end{equation}
where $y_j=1$ is the positive sample, $b_j$ and $\hat{b}$ denotes the $j$-th predicted bounding box and 
the normalized ground-truth bounding box respectively.
$\lambda_{G} =2 $ and $\lambda_{1} = 6$ are used in our settings.

{\noindent \textbf{Post-processing for Online Tracking. }} During online tracking, the size and aspect ratio of the bounding box change slightly across consecutive frames. 
To supervise the prediction using this spatial and temporal consistency, 
similar to SiamCAR~\cite{SiamCAR}, we adopt a scale change penalty $p$ to adjust the classification score map $cls$.
And the final score map $cls'$ can be defined as
\begin{equation}
\begin{split}
\label{equation:window_penalty}
cls' = (1 - w)cls \times p + w \times H
\end{split}, 
\end{equation}
where $H$ is the Hanning window with the shape of $32 \times 32$ and the balance weight $w$.
With the penalty and Hanning window, the confidence of feature points with severe deformations or far from the target in the previous frames will be punished. 
Finally, we select the regressed box corresponding to the highest confidence score as the tracking result.

\subsection{State-of-the-art Comparisons}
We compare our methods with recent state-of-the-art trackers on six tracking benchmarks. The detailed comparison results on the LaSOT~\cite{LaSOT}, TNL2K~\cite{TNL2K}, TrackingNet~\cite{trackingnet}, UAV123~\cite{UAV}, NFS~\cite{NFS} and OTB100~\cite{OTB2015} are reported in Tab.~\ref{tab-sota} and Tab.~\ref{tab-sota-small}. Notably, CNNInMo runs at \textbf{47} {\it fps} on GTX2080Ti GPU, while TransInMo/TransInMo* run at \textbf{67/34} {\it fps} respectively.

{\noindent \textbf{LaSOT}} is a large-scale tracking benchmark containing 1,400 videos. We evaluate different tracking algorithms on its 280 testing videos. As reported in Tab.~\ref{tab-sota}, when applying GIM and InBN to SiamCAR without any other modifications, CNNInMo obtains gains of 3.2/2.9 points on success (SUC) and precision (P) scores, respectively. This evidences the effectiveness of the proposed interaction module and mechanism. Our transformer version with simple cross-correlation matching outperforms recent state of the arts TransT~\cite{TransT} and TrDiMP~\cite{TrDiMP}.

{\noindent \textbf{TNL2K}} is a recently proposed large-scale benchmark for tracking by natural language and bounding box initialization. We evaluate our tracker on its 700 testing videos. CNNInMo outperforms the baseline tracker SiamCAR for 6.9/3.5 points on SUC and precision, respectively.

{\noindent \textbf{UAV123}} contains $123$ aerial sequences captured from a UAV platform. As shown in Tab.~\ref{tab-sota}, the proposed methods (Transformer versions) perform the best among all compared trackers. The CNN version CNNInMo exceeds 1.5 SUC points than the baseline tracker SiamCAR.


{\noindent \textbf{NFS}} consists of 100 challenging videos with fast-moving objects. We evaluate the proposed trackers on the 30 fps version, as presented in Tab.~\ref{tab-sota}. TransInMo and TransInMo* obtain the best two performances on success and precision scores. CNNInMo is also superior to the baseline tracker SiamCAR, which shows that our method can improve the robustness of tracking fast-tracking moving targets.

\begin{table}[!t]
\caption{Results comparison on TrackingNet and OTB100.}
\label{tab-sota-small}
\begin{center}
\resizebox{\linewidth}{!}{
\begin{tabular}{c ccccccccc}
\hline
\specialrule{0em}{1pt}{1pt}
&\textbf{TransInMo*} &TransInMo & CNNInMo &TransT  &SiamCAR \\
\hline
\specialrule{0em}{1pt}{1pt}
\makecell[c]{TrackingNet\\(AUC)}  &\textcolor{red}{81.7} &\textcolor{blue}{81.6} &72.1 &81.4  &65.3   \\
\hline
\specialrule{0em}{1pt}{1pt}
\makecell[c]{OTB100\\(AUC)} &\textcolor{red}{71.1} &\textcolor{blue}{70.6} &70.3 &69.4 &70.0  \\
\specialrule{0em}{1pt}{1pt}
\hline
\end{tabular}}
\end{center}
\end{table}

{\noindent \textbf{TrackingNet and OTB100}} We further evaluate the proposed trackers on TrackingNet and OTB100, as presented in Tab.~\ref{tab-sota-small}. It shows that CNNInMo surpasses the baseline tracker SiamCAR for 6.8 points on TrackingNet. The TransInMo and TransInMo* are on par with (slightly better than) TransT on these two benchmarks.

\subsection{Ablation Study}
\label{subsec-as}

\begin{table}[!t]
  \caption{Component-wise analysis of GIM on TransInMo*.}
  \label{tab-ab}
  \begin{center}
  \resizebox{\linewidth}{!}{
  \begin{tabular}{|c|c|c|c|ccc|c|}
  \hline
  \multirow{2}{*}{\# NUM} &\multirow{2}{*}{Method}  &\multirow{2}{*}{WP}  &\multirow{2}{*}{TCA} &\multicolumn{3}{c|}{LaSOT} &\multirow{2}{*}{FPS} \\
  \cline{5-7}
  & & & &AUC & P$_{Norm}$ &P &\\
  \cline{4-6}
  \hline
  \specialrule{0em}{1pt}{1pt}
  \ding{172}&TransInMo* & &   &61.1	&70.3	&64.9 &30\\
  \specialrule{0em}{1pt}{1pt}
  \ding{173}&TransInMo* &$\surd$ &     &64.6	&74.7	&68.8 &37\\
  \specialrule{0em}{1pt}{1pt}
  \ding{174}&TransInMo* &  &$\surd$     &63.2	&73.6	&68.0 &27\\
  \specialrule{0em}{1pt}{1pt}
  \ding{175}&TransInMo* &$\surd$ &$\surd$  &\textcolor{red}{\textbf{65.7}}	&\textcolor{red}{\textbf{76.0}}	&\textcolor{red}{\textbf{70.7}} &34\\
  \specialrule{0em}{1pt}{1pt}
  \hline
  \end{tabular}}
  \end{center}
  
  \end{table}

\begin{table}[!t]
  \caption{Ablation study on different matching settings.}
  \label{tab-matching}
  \begin{center}
  \resizebox{\linewidth}{!}{
  \begin{tabular}{|c|c|c|c|c|ccc|c|}
  \hline
  \multirow{3}{*}{\# NUM} &\multirow{3}{*}{Method}  &\multirow{3}{*}{\tabincell{c}{DW}}  &\multicolumn{2}{c|}{Matching in TransT}  &\multicolumn{3}{c|}{LaSOT}&\multirow{3}{*}{FPS}\\
  \cline{4-8}
  & & &\tabincell{c}{Encoder\\Layer Num} &\tabincell{c}{Decoder\\Layer Num} &AUC &P$_{Norm}$ &P &\\
  \hline
  \specialrule{0em}{1pt}{1pt}
  \ding{172}&TransInMo &$\surd$ & &   &65.3	&\textcolor{blue}{74.6}	&\textcolor{blue}{69.9} &\textcolor{red}{67}\\
  \specialrule{0em}{1pt}{1pt}
  \ding{173}&TransInMo* & &0 &1 &65.0	&74.3	&69.8 &\textcolor{blue}{59}\\
  \specialrule{0em}{1pt}{1pt}
  \ding{174}&TransInMo* &  &1  &1   &65.3	&73.7	&69.0 &52\\
  \specialrule{0em}{1pt}{1pt}
  \ding{175}&TransInMo* &  &2  &1   &65.5	&73.7	&69.0 &46\\
  \specialrule{0em}{1pt}{1pt}
  \ding{176}&TransInMo* &  &3  &1   &\textcolor{blue}{65.7}	&74.1	&69.2 &40\\
  \specialrule{0em}{1pt}{1pt}
  \ding{177}&TransInMo* & &4 &1  &\textcolor{red}{\textbf{65.7}}	&\textcolor{red}{\textbf{76.0}}	&\textcolor{red}{\textbf{70.7}} &34\\
  \specialrule{0em}{1pt}{1pt}

  \hline
  \end{tabular}}
  \end{center}
  \end{table}
  \label{performance}

{\noindent \textbf{Components in GIM.}}
As shown in Tab.~\ref{tab-ab}, we analyze the influence of WP and TCA in \textbf{GIM} based on \textbf{TransInMo*}. CSA is the basic modeling layer in GIM, hence we only ablate the other two modules.
WP aims at enhancing local perception ability and reducing the complexity of relation learning. Tab.~\ref{tab-ab} shows that without WP and TCA, the tracker obtains an AUC score of 61.1 on LaSOT. When integrating WP into the model, it brings 3.5 points gains (\ding{173} $vs$ \ding{172}) with faster running speed (37 {\it fps} $vs$ 30 {\it fps}).
The TCA module builds branch-wise interaction between the reference and candidate images inside the backbone. As in Tab.~\ref{tab-ab}, TCA brings gains of 2.1 points on AUC (\ding{174}). When applying WP and TCA together, further improvement is obtained with an AUC score of 65.7 (\ding{175}), which proves the rationality of our design.

{\noindent \textbf{Matching Process.}}
We ablate different matching process for our trackers as shown in Tab.~\ref{tab-matching}. With the simple depth-wise correlation (DW) (\ding{172}, \emph{i.e.} TransInMo), the tracker has already achieved compelling performance and speed. We then replace DW with more complex matching modules in TransT~\cite{TransT} with different Transformer encoder numbers. Surprisingly, it (\ding{177}) does not obtain significant gains. As we decrease the encoder layers, the performance is even worse than the simple DW (\ding{172}). Our framework indicates that a tracker with InBN mechanism reduces the demand of fussy matching network to filter target-irrelevant distractors. Matching inside the backbone may be a worth choice in future work.
 
 \begin{figure}[!t]
\begin{minipage}{0.99\linewidth}
\centerline{\includegraphics[width=0.99\textwidth]{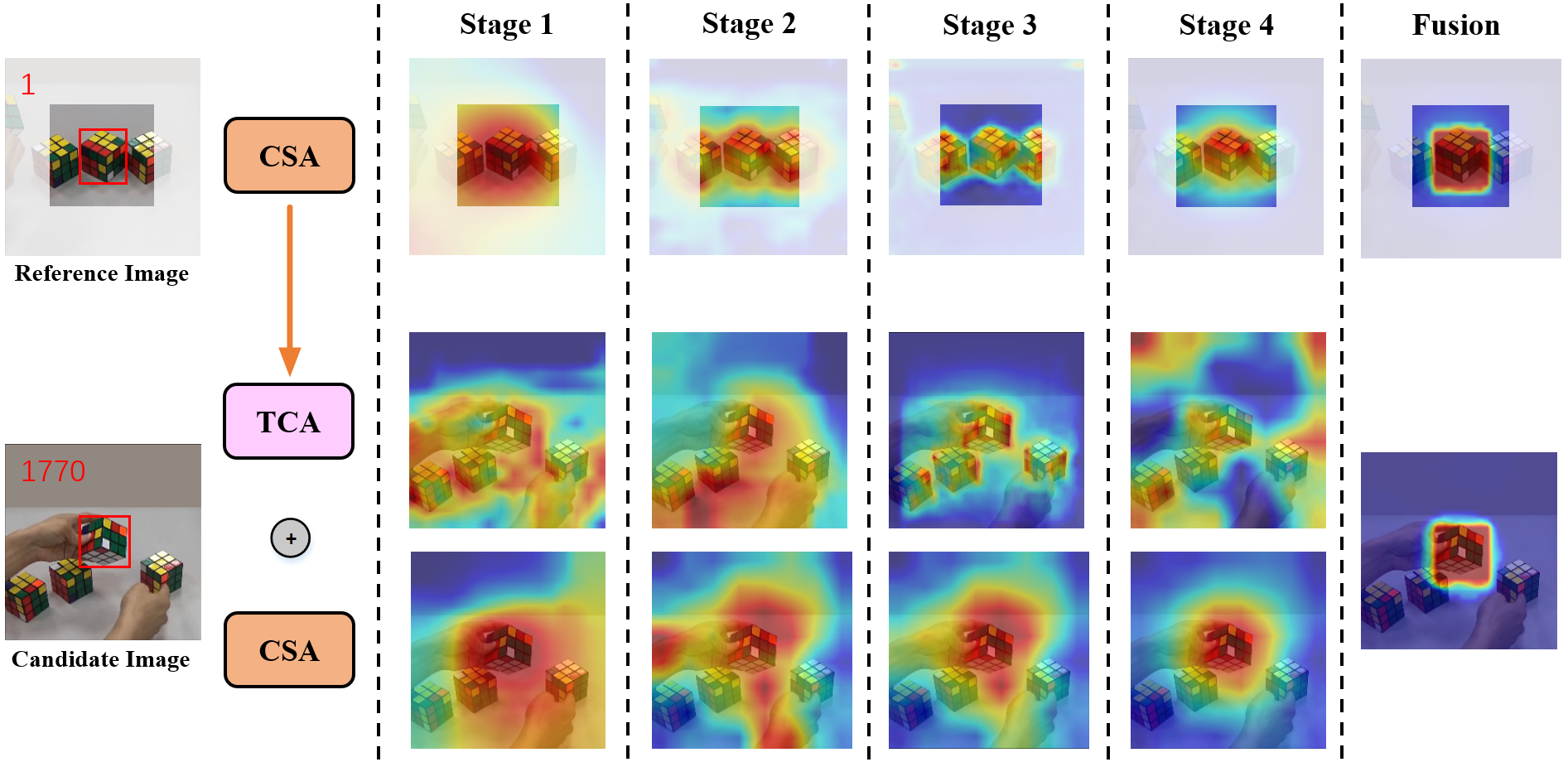}}
\end{minipage}
\caption{Activation maps for CSA and TCA.}
\label{fig:cam}
\end{figure}

\begin{figure}[!t]
\begin{minipage}{0.99\linewidth}
\centerline{\includegraphics[width=\textwidth]{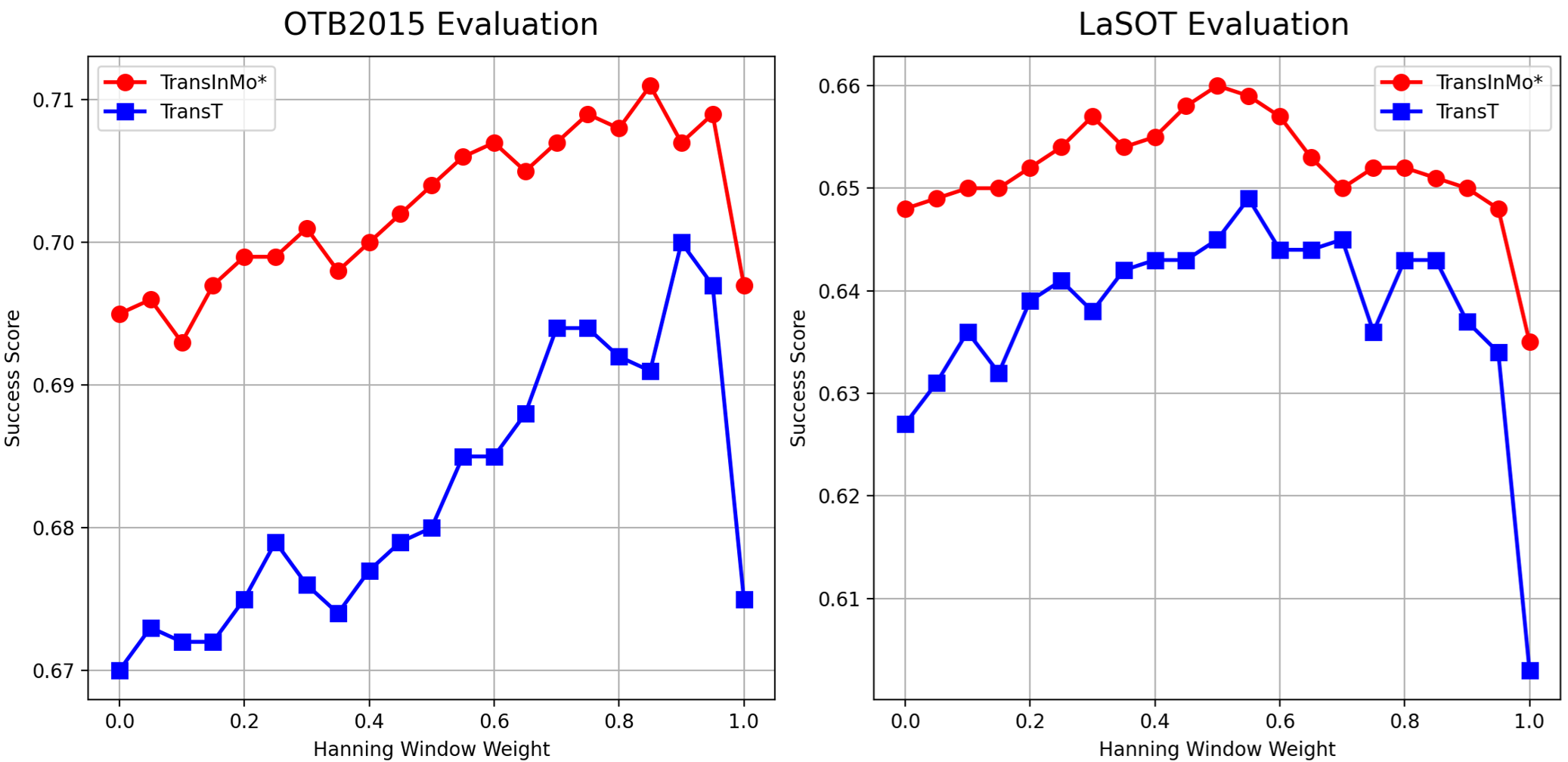}}
\end{minipage}
\caption{The influence of the Hanning window.}
\label{fig:win}
\end{figure}

{\noindent \textbf{Activation Map Visualization.}} To better understand our CSA and TCA, we visualize their activation maps in Fig.~\ref{fig:cam}. As shown, when the target appearance changes heavily or similar distractors exist, the branch-wise interaction of TCA can help the backbone sense the target-relevant areas. Then self-attention of CSA employs the context messages to filter useless messages and focus on the target area.

{\noindent \textbf{Hanning Window Influence.}} In Fig.~\ref{fig:win}, we present the influence of post-processing (i.e. Hanning window) used in our TransInMo* and TransT on two benchmarks. From Fig.~\ref{fig:win}, we observe that the proposed TransInMo* is more stable. In the case without post-processing (\emph{i.e.} weight=0), our tracker still shows compelling performance, which indicates the effectiveness of feature representation learned by our backbone.

\begin{figure}[!t]
  \begin{minipage}{0.99\linewidth}
  \centerline{\includegraphics[width=\textwidth]{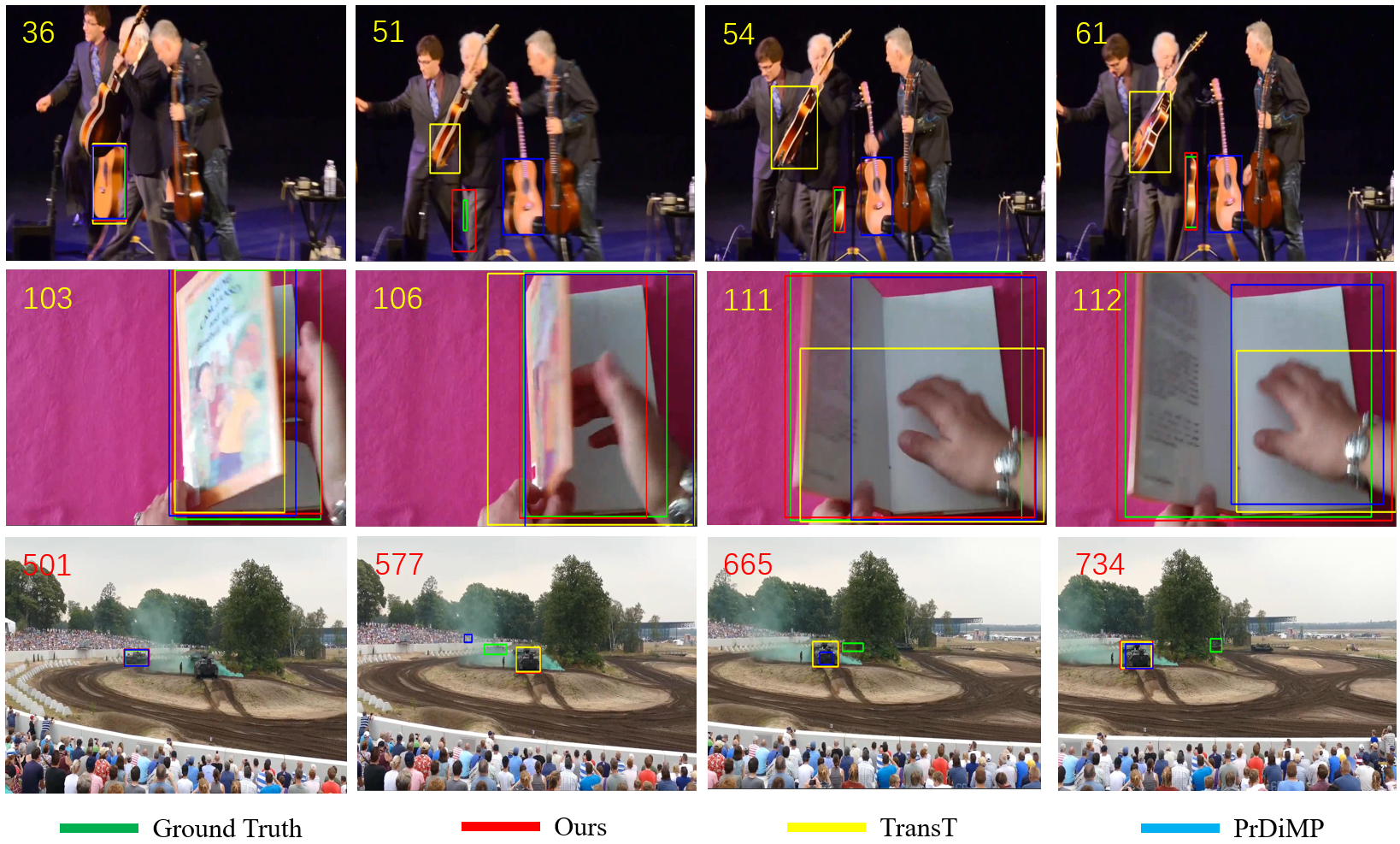}}
  \end{minipage}
   
  \caption{Results visualization of different trackers. The first two rows show that \textbf{TransInMo*} achieves the best performances. 
  The last row demonstrates failure cases of our tracker.}
  \label{fig:vis}
  \end{figure}

{\noindent \textbf{Visualization and Failure Cases.}}
As shown in Fig.~\ref{fig:vis}, our TransInMo* delivers accurate tracking under deformation (the first row) and large scale variation (the second row). It demonstrates the resilience of our tracker and effectiveness of GIM and InBN in complex environments. In the third row, we also show the failure case of our tracker. In this case, the target is fully occluded for about 200 frames, causing our tracker could not get helpful messages from the reference representation and surrounding context.
One possible solution for this circumstance is using global search when the target disappears for a long time. But this is beyond the topic of our paper. We leave this to future study.


\section{Conclusions}
This work presents a novel mechanism that conducts branch-wise interactions inside the visual tracking backbone network (InBN) via the proposed general interaction modeler (GIM). The core idea is injecting target information into the representation learning of candidate frame and providing direct feedback to backbone network inference. We prove that both the CNN network and the Transformer backbone can enjoy bonus brought by InBN. The backbone network can build more robust feature representation under complex environments with the improved target-perception ability. Our method achieves compelling tracking performance by applying the backbones to Siamese tracking.

\clearpage

\bibliographystyle{named}
\bibliography{ijcai22}

\end{document}